# High-order Graph Neural Networks with Common Neighbor Awareness for Link Prediction

Ling Wang
*School of Computer Science and Technology*
*Chongqing University of Posts and Telecommunications*
Chongqing, China
d220201030@stu.cqupt.edu.cn

Minglian Han
*College of Computer and Information Science*
*Southwest University*
Chongqing, China
hanmlian@email.swu.edu.cn

*Abstract*—Link prediction is a fundamental task in dynamic graph learning (DGL), inherently shaped by the topology of the DG. Recent advancements in dynamic graph neural networks (DGNN), primarily by modeling the relationships among nodes via a message passing scheme, have significantly improved link prediction performance. However, DGNNs heavily rely on the pairwise node interactions, which neglect the common neighbor interaction in DGL. To address this limitation, we propose a High-order Graph Neural Networks with Common Neighbor Awareness (HGNN-CNA) for link prediction with two-fold ideas: a) estimating correlation score by considering multi-hop common neighbors for capturing the complex interaction between nodes; b) fusing the correlation into the message-passing process to consider common neighbor interaction directly in DGL. Experimental results on three real DGs demonstrate that the proposed HGNN-CNA acquires a significant accuracy gain over several state-of-the-art models on the link prediction task.

## I. INTRODUCTION

DYNAMIC graphs (DG) are prevalent in various real-world scenarios, including social networks [1-10], e-commerce platforms [11-18], and biological systems [21-24]. Link prediction on DG plays a crucial role in numerous applications, such as recommendation systems [14] and drug discovery [22], and primarily depends on effective DGL [24-28]. A key challenge in designing a successful learning framework for DG lies in developing efficient mechanisms to capture the intricate relationships among nodes (e.g., structural correlation). This mechanism ensures that nodes with close relationships are represented by similar embedding vectors, thereby enhancing the quality of DGL and the performance of the link prediction task.

Dynamic Graph Neural Networks (DGNNs) have shown remarkable performance in DGL [29-48]. In various DGNNs, neighbor aggregation is a key common ingredient, which updates node representations by weighted aggregating information from nodes' neighbors via a message passing scheme [18, 25, 29, 41]. For instance, WDGCN [43] and EvolveGCN [47] that adopt a spectral-based filter, where each node generates the representation by aggregating features from its neighbors. However, the aforementioned methods only adopted averaging weights to implement message-passing, so they fail to make full advantage of the rich information in node features. Hence, other studies [48-56] measure the node feature correlations for message passing process. For instance, DySAT [54] learns an additional feature correlation score so that important neighbors are given a higher weight in message passing. Though effective, existing neighbor aggregation in DGNNs has two challenges.

*Firstly*, DGNNs' aggregating weight pays more attention to links or node features, but less attention to graph-structure information, which is critical for link prediction. Recently, heuristic algorithms [7, 57-73] that only leverage graph-structure information obtain competitive performance in link prediction. However, in existing DGNNs, graph-structure information is used solely to determine neighboring nodes during the message passing and neighbor aggregation process. Hence, the graph-structure information has remained underexplored to be incorporated into message passing in DGNNs for link prediction.

*Secondly*, the existing DGNNs often overlook the common neighbor complex interaction. Most DGNNs assume that the interaction relationships of nodes are purely pairwise, so correlation calculations are only performed between the connected pairwise nodes. However, this assumption is inconsistent with real-world scenarios involving common interactions of nodes. The common neighbors in the interaction could affect their correlation. The potential benefit of considering common interactions in DGNNs is illustrated with a simple example in Fig. 1. According to common neighbor properties [74], in graph snapshot $G_t$, node

---

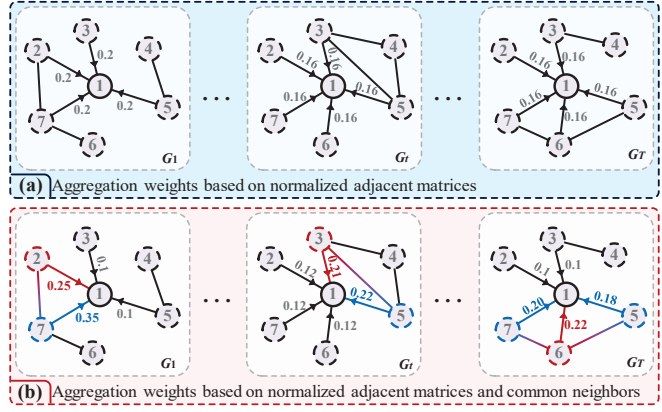

Fig. 1. Example of a DG with aggregation weights between nodes for spatial message passing. The aggregation weights in (a) are from normalization of the adjacent matrices without considering common neighbor interactions. The weights in (b) are calculated by considering both adjacency matrix and the common neighbor (marked by a blue/red dashed circle in (b)), so the refined aggregation weights for the nodes are different than those in (a).

1 should have a high aggregation weight to node 2, as they have a common neighbor 7. However, without considering the common neighbor information in Fig. 1(a), the aggregation weight between nodes 1 and 3 is equal to that between nodes 1 and 3, as only considering the connected information between nodes. Conversely, by considering common neighbor information as shown in Fig. 1(b), the aggregation weight between nodes with common neighbors is increased.

To address the aforementioned challenges, this study proposes a **H**igh-order **G**raph **N**eural **N**etwork with **C**ommon **N**eighbor **A**wareness (HGNN-CNA), specifically designed for link prediction on DGs. HGNN-CNA aims to measure common neighbor correlation via graph-structure information, which is incorporated into the message passing process. Furthermore, it exploits high-order graph neural networks (HGNN) [77-80] to perform message passing to obtain a DG representation for link prediction.

The main contributions of this study can be summarized as:

a) **Common neighbors are captured**. Structural features are learned to estimate common neighbor correlation scores via capturing common neighbors in DG; and

b) **Aggregation weights are refined.** Common neighbor correlation score is fused with aggregation weights in message passing to obtain the refined aggregation weights.

## II. PRELIMINARIES

In this section, we first present the downstream task to be solved by learning DG representations. Then, we introduce the necessary preliminaries on HGNN.

### A. Problem Formulation

In this study, a DG is defined as $G = \{G_t\}_{t=1}^{T}$, and each graph snapshot $G_t = (V, E_t, X_t)$ is a graph at time slot $t$ ($1 \le t \le T$), where $V$ is the node-set whose number of nodes is $N=|V|$. $E_t$ represents the edge set at time slot $t$, and $X_t$ contains all node feature vectors. For a graph snapshot $G_t$, $E_t$ can be represented as an adjacency matrix $A_t \in \mathbb{R}^{N \times N}$. Hence, a DG can also be described by an adjacency tensor $\mathbf{A}=[a_{ijt}] \in \mathbb{R}^{N \times N \times T}$ and a feature tensor $\mathbf{X}=[x_{ijt}] \in \mathbb{R}^{N \times F \times T}$ naturally. Given a DG, the goal of this study is to obtain a node embedding tensor $\mathbf{H}=[h_{it}] \in \mathbb{R}^{N \times F \times T}$ by capturing spatial-temporal patterns, and then the link between nodes is predicted via node embedding $\mathbf{H}$.

### B. High-order Graph Neural Network

High-order Graph Neural Network (HGNN) built on the tensor product is a straightforward and efficient DGNN for learning DG. In our framework, DG is represented using tensors, and HGNN is employed as the fundamental model to learn DG for the link prediction task. The following section introduces the tensor product and HGNN.

**Tensor product**. Given two three-order tensors $\mathbf{X} \in \mathbb{R}^{I \times J \times T}$ and $\mathbf{Y} \in \mathbb{R}^{J \times K \times T}$, and an invertible matrix $M \in \mathbb{R}^{T \times T}$, the tensor product is denoted by $\mathbf{X} \circledast \mathbf{Y} \in \mathbb{R}^{I \times K \times T}$ and is defined as:

$$\mathbf{X} \circledast \mathbf{Y} = \left( \left( \mathbf{X} \times_3 M \right) \otimes \left( \mathbf{Y} \times_3 M \right) \right) \times_3 M^{-1}, \tag{1}$$

where $\times_3$ is tensor mode-3 product for information mixing and $\otimes$ denotes tensor face-wise product [75, 76].

**HGNN**. According to [77-80], High-order Graph Neural Network (HGNN) based on the tensor product is defined as:

$$\begin{cases} \mathbf{H}^{(0)} = \mathbf{X}, \\ \mathbf{H}^{(l)} = \sigma\left( \hat{\mathbf{A}} \circledast \mathbf{H}^{(l-1)} \circledast \mathbf{W}^{(l)} \right), \end{cases} \tag{2}$$

where $\sigma$ denotes a non-linear activation function like Sigmoid, $\mathbf{X}$ is the node feature tensor, $\mathbf{W}^{(l)} \in \mathbb{R}^{F \times F \times T}$ is a learnable weight tensor for feature transformation, and $\mathbf{H}^{(l)} \in \mathbb{R}^{N \times F \times T}$ is the node representation tensor. $\hat{\mathbf{A}}$ is the normalization adjacent tensor.

## III. PROPOSED MODEL

In this paper, we propose a high-order graph neural network with common neighbor awareness for link prediction. The network consists of two main components as follows: a) *Structural Feature Learning Module*, which is adopted to embed edge information into node-level structural feature tensor. b) *Common Neighbor Awareness Module*, which exploits common neighbor's structural features to obtain common neighbor correlations, which are fused into aggregation weights for message passing. The details of each module are presented in the following.

### A. Structural Feature Learning Module

In this study, a structural feature learning module is designed to learn node-level structural features for capturing common neighbors. As illustrated in existing works [7, 81-84], to capture common neighbors, a potential technique is to adopt exponentiation of the adjacency tensor $\mathbf{A}$. However, it only uses predefined edge information, which may fail to capture complex and implicit structural information. For this issue, HGNN-CNA consists of a structural feature generator to generalize and learn these structural features. Specifically, a structural feature generator $g_\theta$ is adopted to generate structural features of each node using only the adjacent matrix $A_t$ of DG as:

$$s_{it} = g_\theta(A_t) = g_{node}\left(\sum_{j \in \mathcal{N}(i)} g_{edge}(a_{ijt})\right), \quad (3)$$

where $s_{it}$ is a structural feature value of the node $i$ at time slot $t$ and $g_\theta$ is a learnable function comprised of MLPs, $g_{node}$ and $g_{edge}$, for nodes and edges, respectively. In this way, HGNN-CNA takes only an adjacency tensor $\mathbf{A}$ as an input and learns to generate a beneficial node-level structural feature tensor $\mathbf{S}=[S_t]\in\mathbb{R}^{N\times 1\times T}$ from DG adaptively.

Conventional DGNNs cannot directly capture common neighbors for two reasons: the lower dimension of hidden embeddings compared to the number of nodes (i.e., $F \ll N$) and the normalized adjacency matrix. The former makes the neighborhoods indistinguishable after aggregation, which hinders the detection of common neighbors. Further, the latter prevents DGNNs from counting the number of neighborhoods. To address the first issue, the obtained structural feature vectors at each time slot $t$ are constructed as a diagonal structural feature matrix $S_t \in \mathbb{R}^{N \times N}$:

$$S_t = diag(s_{1t}, s_{2t}, s_{3t}, ..., s_{Nt}), \quad (4)$$

where $diag(\cdot)$ turns $z \in \mathbb{R}^d$ into a diagonal matrix of size $n \times n$ that has the values of $n$ on its main diagonal. By doing so, a structural feature tensor $\mathbf{S}=[S_t]\in\mathbb{R}^{N\times N\times T}$ is obtained.

To count the number of common neighbors directly, the unnormalized tensor $\mathbf{A}$ is employed in the procedure to account for the number of common neighbors. Additionally, the procedure adopts a face-wise product to enable common neighbors with structural features:

$$\mathbf{Z} = \mathbf{A} \otimes \mathbf{S}. \quad (5)$$

It should be pointed out that each $i$-th row vector at time slot $t$ of $\mathbf{Z}$, i.e., $z_{it}$, involves all the features of node $i$'s neighboring nodes at time slot $t$ individually. Hence, when the inner product of two vectors at the same time slot in $\mathbf{Z}$ is performed, the scores are obtained via only the common neighbors' features. Specifically, the score is equal to the sum of the squared structural feature value:

$$z_{it} z_{jt}^T = \sum_{k=1}^{N} z_{ikt} z_{jkt} = \sum_{k \in N_t(i) \cap N_t(j)} (s_{kt})^2, \quad (6)$$

where $N_t(i)$ denotes the neighbor set of the $i$-th node at time slot $t$. Further, to account for multi-hop common neighbors, (6) is extended to a multi-hop setting as follows:

$$\mathbf{Z} = \sum_{k=1}^{K} \beta_k \mathbf{A}^k \otimes \mathbf{S}. \quad (7)$$

For clarity, the tensor form of (5) can be presented in matrix form, in which each front slice $Z_t$ in $\mathbf{Z}$ is obtained as:

$$Z_t = \sum_{k=1}^{K} \beta_k A_t^k S_t, \quad (8)$$

where $\beta_k$ represents the weight associated with the $k$-th hop and $K$ decides the maximum number of hops. $\beta_k$ facilitates the integration of multi-hop structural information and controls the relative importance of close versus distant neighbors. $A^k$ is responsible for capturing $k$-hop common neighbors.

### B. Common Neighbor Awareness Module

The common neighbor awareness module aims to estimate the correlation between nodes via common neighbor and fuse the correlation into message passing. To consider the impact of the common neighbor in message passing, the common neighbor correlation between the $i$-th node and the $j$-th node at time slot $t$ is calculated by:

$$\hat{c}_{ijt} = z_{it} z_{jt}^\top, \quad (9)$$

where $\hat{c}_{ijt}$ attempts to obtain the correlation scores about the common neighbor. Then, all common neighbor scores of all node pairs at all time slots can be obtained as a scores tensor $\hat{\mathbf{C}}$.

In the DGNNs' message-passing process, the aggregation weight calculation only considers the information of directly connected node pairs. Hence, it overlooks the influence of common neighbors between nodes. For this issue, we learn the common neighbor interventions. Given the scores $\hat{c}_{ijt}$, we obtain the normalized correlation as:

TABLE I. Experimental Data Statistics

| No. | Dataset | Nodes | Edges | Time slots |
|---|---|---|---|---|
| D1 | math-overflow [62] | 2,598 | 227,706 | 85 |
| D2 | ask-ubuntu [62] | 3,748 | 159,817 | 73 |
| D3 | bitcoin-alpha [63] | 3,783 | 24,187 | 32 |

$$c_{ijt} = \xi(\hat{c}_{ijt}) = \frac{\exp(\hat{c}_{ijt})}{\sum_{k \in N_t(i)} \exp(\hat{c}_{ikt})}, \quad (10)$$

where $\xi$ denotes a SoftMax function that scales the scores, and $c_{ij}$ is the common neighbor correlation.

To effectively consider the common neighbor correlation into the message-passing process, the correlation and aggregation weights are fused to obtain refined aggregation weights. To this end, two parameters, $r_{ct}$ and $r_{at}$ are introduced to fuse the correlations and aggregation weights as:

$$o_{ijt} = r_{ct} \cdot c_{ijt} + r_{at} \cdot \hat{a}_{ijt}, \quad (11)$$

where $r_{ct}$ and $r_{at}$ represent fusing weights. $\hat{a}_{ijt}$ is the aggregation weight from tensor $\hat{\mathbf{A}}$. $o_{ijt}$ is the refined aggregation weight by considering the common neighbor intervention for message passing. Hence, we can obtain a refined aggregation weight tensor $\mathbf{O}$. Then, the refined aggregation weight tensor $\mathbf{O}$ is incorporated into the high-order graph neural network to facilitate message passing, enabling node embedding learning for link prediction.

## C. Model Optimization

This study learns a DG representation for link prediction. For the link between node $i$ and $j$ at time slot $t$, two node embedding vectors are concatenated and fed into an MLP to obtain the link probability as:

$$\hat{y}_{ijt} = \rho([h_{it} \| h_{jt}]), \quad (12)$$

where $\hat{y}_{ijt}$ is the link probability on the concatenation of the above two node embedding vectors. $\rho$ is an MLP to obtain a link probability value, and $[\cdot\|\cdot]$ denotes the concatenation operation of node embedding vectors.

To optimize model parameters, a learning objective is built:

$$L = \frac{1}{|\Lambda|} \sum_{y_{ijt} \in \Lambda} \left( y_{ijt} \log \hat{y}_{ijt} + (1 - y_{ijt}) \log(1 - \hat{y}_{ijt}) \right) + \alpha \|\Omega\|_2, \quad (13)$$

where $L$ represents the loss of the objective function. $y_{ijt}$ is the link label between node $i$ and $j$ at time slot $t$. $\Omega$ represents the parameter set in the model to optimize, and $\Lambda$ is the training set. $\|\cdot\|_2$ is $L_2$ regularization with coefficient value $\alpha$.

## IV. EXPERIMENT AND RESULTS

In this section, we present three dynamic graph datasets to evaluate the proposed model HGNN-CNA, and compare it with existing state-of-the-art models.

### A. Experiment Settings

**Datasets.** In this study, three DG datasets in Table I are adopted, and each dataset is split by the ratio of 70%, 20%, and 10% as training set, validation set, and test set, respectively. Similar to [54], negative sampling is conducted.

**Evaluation Protocols.** In this study, for the link prediction task in DG representation learning, F1-scores [47, 75] and Accuracy [54, 76] are adopted to evaluate the proposed model.

TABLE II. Comparison results on F1-Scores and Accuracy. The best results are in **bold**, and the suboptimal results are underlined for clarity.

| Models | Metrics | D1 | D2 | D3 |
|---|---|---|---|---|
| M1 | F1-Score | 0.7755 | 0.7641 | 0.7220 |
|    | Accuracy | 0.7486 | 0.7300 | 0.6962 |
| M2 | F1-Score | 0.7667 | 0.7519 | 0.7300 |
|    | Accuracy | 0.7348 | 0.7173 | 0.7039 |
| M3 | F1-Score | 0.8064 | 0.7913 | 0.7844 |
|    | Accuracy | 0.7603 | 0.7676 | 0.7553 |
| M4 | F1-Score | 0.8123 | 0.8069 | 0.7901 |
|    | Accuracy | 0.7885 | 0.7784 | 0.7695 |
| M5 | F1-Score | 0.8239 | 0.8166 | 0.8093 |
|    | Accuracy | 0.8066 | 0.7983 | 0.7846 |
| M6 | F1-Score | <u>0.8325</u> | <u>0.8247</u> | <u>0.8134</u> |
|    | Accuracy | <u>0.8164</u> | <u>0.8002</u> | <u>0.7935</u> |
| M7 | F1-Score | **0.8496** | **0.8438** | **0.8357** |
|    | Accuracy | **0.8285** | **0.8163** | **0.8114** |

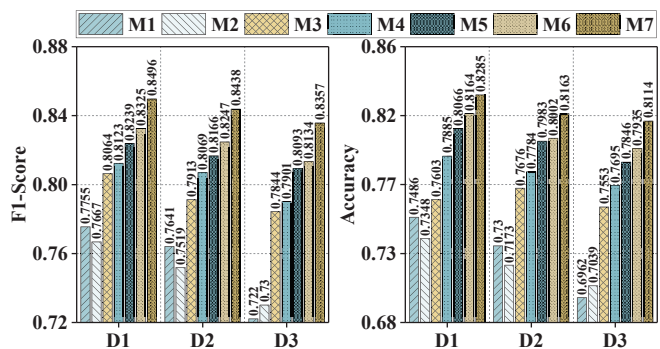

Fig. 2. F1-Score and Accuracy for link prediction.

*Evaluation Protocols.* In this study, for the link prediction task in DG representation learning, F1-scores [47] and Accuracy [48] are adopted to evaluate the proposed model.

*Compared Models.* To evaluate the proposed HGNN--CNA model, we compare it with six baseline DG learning models, which include two static graph convolution models (i.e., GCN [61] (M1), LightGCN [42] (M2)), four dynamic graph convolution models (i.e., WDGCN [43] (M3)), EvolveGCN [47] (M4), DySAT [54] (M5), FSTGCN [78] (M6)). In addition, M7 denotes the proposed HGNN-CNA.

*Training Settings.* To ensure a fair comparison, the following training setting is adopted:

a) The HGNN-CNA is implemented in the PyTorch framework with four 2080Ti GPU cards;
b) For all comparison models, the Adam optimizer is used, with the learning rate tuned from the set {0.1, 0.01, 0.02, 0.05, 0.001, 0.002}. $L_2$ regularization's coefficient value $\alpha$ is tuned from the set {0.01, 0.005, 0.001, 0.0005}, and the layer of HGNN-CNA is set as 2, i.e., $l$=2, and $r_{ct}$=$r_{at}$=0.5. The feature dimension is set to 32, and hyperparameters are carefully tuned to obtain optimal performance;
c) The termination conditions are defined as follows: the training process halts if the iteration threshold reaches 300 or if no improvement in performance is observed for 10 consecutive iterations.

### B. Comparison with State-of-the-Art Models

To validate the link prediction effectiveness of the proposed HGNN-CNA, we compare it with six peer models. Table II and Fig. 2 show the F1-Score and Accuracy obtained by all the above models on the testing set. From the result, we can have the following findings.

a) Compared with two static graph neural networks (M1 and M2), M7 obtains excellent performance in DG representation learning for link prediction. For instance, according to Table II on D1, M7's F1-Score is 0.8496, higher than 0.7755 and 0.7667 obtained by M1 and M2, respectively. Similar outcomes are obtained in the other datasets as shown in Table II. Intuitively, the main reason is that M1 and M2, as static graph neural networks, neglect the temporal patterns in DG, leading to inferior DG representation learning performance for link prediction.
b) M7 also obtains notable performance gain compared with four dynamic GNN models, i.e., M3-M6, which adopt a simple normalized adjacent matrix or node feature correlations between nodes for message passing. For instance, as shown in Table II on D1, M7's F1-Score is 0.8496, which is higher than M3-M6. On the evaluation metric Accuracy, M7 obtains 0.8285, which is also better than M3-M6. Similar results can be found in other testing datasets. Intuitively, the above model ignores the influence of common neighbors of nodes, while HGNN-CNA can obtain more accurate node embedding and improve the accuracy of link prediction by capturing common neighbor correlation for message passing.

From the above experiment analysis, it is verified that the proposed HGNN-CNA can improve the performance of link prediction via common neighbor awareness.

## V. CONCLUSION

In this paper, we propose a High-order Graph Neural Network with Common Neighbor Awareness (HGNN-CNA) for link prediction. HGNN-CNA learns the structural features and calculates node correlations by common neighbor features. These correlations are then incorporated into the aggregation weights during message passing, enabling the model to learn more expressive node embeddings for link prediction. Furthermore, we conduct comparative experiments on DG datasets, demonstrating that common neighbor correlation enhances link prediction performance. In future work, we plan to explore the impact of common neighbor correlations on different message passing schemes.